\documentclass[lettersize,journal]{IEEEtran}
\usepackage{amsmath,amsfonts}
\usepackage{algorithmic}
\usepackage{algorithm}
\usepackage{array}
\usepackage{booktabs}
\usepackage[table]{xcolor} 
\usepackage{textcomp}
\usepackage{stfloats}
\usepackage{url}
\usepackage{verbatim}
\usepackage{graphicx}
\usepackage{caption}
\usepackage{subcaption}
\usepackage{cite}
\usepackage{xr}
\usepackage[colorlinks=true, urlcolor=blue]{hyperref}

\begin{document}

\title{Channel-Level Relation to Attentive Aggregation with Neighborhood-Homogeneity Constraint  for Point Cloud Analysis}
\author{
\IEEEauthorblockN{%
 Jiaqi Shi\textsuperscript{1},
 Jin Xiao*\textsuperscript{1},
 Xiaoguang Hu\textsuperscript{1},
 Wenxuan Ji\textsuperscript{1}, \\
 Zichong Jia\textsuperscript{1}, 
 Zifan Long\textsuperscript{1},
 Tianyou Chen\textsuperscript{2},
 Baochang Zhang\textsuperscript{3}\\
}
\IEEEauthorblockA{
\textsuperscript{1}School of Automation Science and Electrical Engineering, Beihang University, Beijing, China \\
\textsuperscript{2}Wuhan Leaddo Measuring \& Control Technology, Wuhan, China \\
\textsuperscript{3}School of Artificial Intelligence, Beihang University, Beijing, China\\
Emails: shijiaqi@buaa.edu.cn, xiaojin@buaa.edu.cn, xiaoguang@buaa.edu.cn, jiwenxuan@buaa.edu.cn, jiazichong@buaa.edu.cn, longzifan@buaa.edu.cn, chentianyou@buaa.edu.cn, bczhang@buaa.edu.cn
}

}

\maketitle



\begin{abstract}
In 3D point cloud understanding, the core challenge lies in accurately capturing discriminative features within complex neighborhoods, which directly affects the execution precision of downstream tasks such as embodied AI and autonomous driving. Existing methods explore feature correlation discrimination but are limited to point-level spatial distribution or channel responses, enabling only coarse-grained level evaluation. For modern multi-scale point cloud networks, such coarse-grained metrics inevitably incur significant information loss in deeper layers.
To address this, we propose PointCRA, a novel network with a channel-level metric-based enhancement mechanism. Our core idea is to introduce temporal trend variation as a new evaluation dimension to avoid the information loss caused by weight dimension collapse in existing spatial and channel attention mechanisms. On this basis, we construct a multi-level calibration framework guided by neighborhood homogeneity for weight calibration, and design a dedicated loss function to enhance channel discriminability.PointCRA leverages intrinsic feature priors to adaptively correct feature aggregation, offering interpretability with low parameter overhead.
Our method is transferable, interpretable, and efficient. We validate the proposed method on diverse datasets and benchmark models, and further demonstrate its rationality through extensive analytical experiments. Our PointCRA achieves 77.5\% mIoU on the S3DIS dataset, 90.4\% OA on the ScanObjectNN dataset, and 87.4\% instance mIoU on the ShapeNetPart dataset. The code and pretrained weights are publicly available on GitHub: \url{https://github.com/AGENT9717/PointCRA}.

\end{abstract}

\begin{IEEEkeywords}
Point Cloud, Deep Learning, Neighborhood Aggregation, Attention.
\end{IEEEkeywords}

\section{Introduction}
\IEEEPARstart{P}{oint} clouds are a key representation of real-world 3D geometry, essential for perception tasks like embodied AI and autonomous driving. Their unstructured and unordered nature makes the accurate extraction of discriminative features from local neighborhoods a core challenge\cite{pointgt,zhu2025sgg,Dupmam}.

The pioneering work PointNet\cite{pointnet} first successfully introduced deep learning to point cloud analysis by independently mapping the coordinates of each point into a feature vector, thereby establishing the architectural foundation for point cloud deep learning.
Subsequent researches\cite{pointnext,pointmeta,pointvector}  primarily follow the path of architectural deepening, stacking multiple neighborhood aggregation modules to enhance representational capacity in complex scenes.
However, their performance  is constrained by the inherent randomness of group-sampling method in neighborhood aggregation (e.g., ball query or k-nearest neighbors), which relies solely on simple spatial heuristics like a fixed radius or a predefined number of points. This often introduces irrelevant and noisy features into neighborhoods, interfering with aggregation. Given that the generalization ability of deep features is negatively correlated with their Gaussian complexity \cite{pointconvformer,rademacher,filter} such noise interference inevitably degrades network performance.

To address this issue, recent works shift focus from architectural scaling to refining the aggregation process itself. Representative approaches include modeling inter-point feature correlations via attention mechanisms\cite{pct,pointtransformerv1,pointtransformerv2,pointtransformerv3}, refining point-to-center affinity with additional spatial positional encoding\cite{x-3d,pointdistribution,pointhop}, or generating discriminative weights by computing channel-wise responses~\cite{canet,overlapping}. These methods perform weighted computation based on the aggregated neighborhood feature matrices, with their metrics confined to a single dimension of the 2D matrix (spatial or channels).

However, such an approach leads to weight dimension collapse, enabling only point-level metric weights. This collapse results in two critical limitations:
(i) Coarse Metric Granularity. These methods treat each neighboring point as an indivisible whole, using its full feature vector for affinity computation. However, high-level points in deep networks are superpoints aggregated from lower-level points, encoding rich substructure. Compressing such complex entities into a single vector inevitably averages out substructure semantics and channel-wise discriminability, leading to fine-grained information loss and aggregation ambiguity.

(ii) Rigid metric perspective. These methods rely entirely on metric references based on query points or global means, making the enhancement effect highly sensitive to the representativeness of such references. However, this representativeness varies significantly across geometric contexts: in homogeneous regions (e.g., walls, tabletops) versus heterogeneous regions (e.g., edges, corners), neither the centrality of a point nor the discriminability of feature channels can be uniformly treated. Continuing to use these fixed references as the sole anchors for weight assignment introduces evaluation bias, resulting in structural distortion in complex regions.

Therefore, we propose PointCRA, a point cloud analysis network that performs adaptive aggregation based on channel-level metrics, transcending the metric granularity limitations of existing methods by introducing a new evaluation dimension.
First, we refine the fundamental unit of affinity computation from the point level to the channel level. Specifically, we introduce a channel-level similarity evaluation based on feature transformation trend distance, enabling fine-grained characterization of each channel across neighboring points.
Second, we construct a channel-point-neighborhood three-level framework for progressive metric and reverse calibration. The framework first performs stepwise aggregation to generate weights and neighborhood homogeneity distributions. It then leverages the homogeneity distribution as a constraint to reverse-calibrate point-level and channel-level weights in a top-down manner. Finally, we design a dedicated loss function to enhance neighborhood weight discriminability, avoiding weight oversaturation.

We transplant PointCRA to multiple baseline variants. Experimental results show that our method serves as a general-purpose enhancement for point cloud feature representation.
The main contributions of this work are summarized as follows:
\begin{itemize}
\item We propose a channel-level affinity metric that captures the correlation strength between neighboring points at the feature channel level, overcoming the limitation of existing methods that operate at the point level, and incorporate lightweight improvements.

\item We construct a three-level weighting framework constrained by neighborhood homogeneity, which enables adaptive weight allocation through channel-to-point-to-neighborhood progressive metric and top-down reverse calibration.

\item we integrate the aforementioned algorithmic designs to propose the PointCRA network, along with a dedicated loss function to enhance weight discriminability. Equipped with these components, PointCRA achieves state-of-the-art performance.

\item We transfer our proposed method to different baseline models and datasets for evaluation, validating its effectiveness and transferability, and conduct comprehensive analytical experiments to verify its rationality.
\end{itemize}

\section{Related Work}
\subsection{Point Cloud Analysis}
Due to the unordered nature of point clouds, traditional image methods cannot be directly applied. PointNet \cite{pointnet} introduced a symmetric function to aggregate global features but struggled to capture fine-grained local structures. To address this, PointNet++ \cite{pointnet++} proposed the Set Abstraction module and established a hierarchical feature learning paradigm, laying the foundation for modern point cloud analysis networks.

In recent years, driven by the demand for complex scene understanding, numerous works have focused on modernizing the PointNet++ architecture to enhance its representation capacity. On one hand, some works \cite{pointnext,pointmeta,pointconv,pointvector,pointnat} stack multiple neighbor aggregation modules within the same layer, constructing deeper and wider networks. On the other hand, other works \cite{pointcnn,pointconv,pointnat,kpconv,kpconvx} optimize neighbor relationship modeling by introducing finer spatial encoding or more advanced aggregation mechanisms, enhancing local feature discriminability while preserving the hierarchical structure.

However, the core aggregation strategy of these methods remains similar to that of PointNet++: updating a central point by aggregating features from its neighbors. When confronting complex scenes, this paradigm still faces two critical challenges: interference from irrelevant points and sub-structure confusion. When a local neighborhood contains points from different objects or semantic parts, simple aggregation introduces noise, leading to feature ambiguity and limiting the model's perception of fine-grained local geometry.

To address these issues, we propose a feature enhancement method tailored for modern multi-scale point cloud analysis networks, which calibrates neighborhood feature consistency by aggregating outputs from multiple aggregation modules, thereby fully unleashing the potential of existing architectures and enhancing feature representation in complex scenarios.

\subsection{Attentive Neighborhood Aggregation}
To address irrelevant point interference in mainstream aggregation methods that rely on 3D coordinates for neighborhood relations, subsequent works have proposed weighted aggregation strategies. These include evaluating neighbor importance via feature similarity \cite{pointconvformer}, incorporating spatial encoding to refine metric similarity \cite{x-3d,pointhop}, adapting weights via learnable kernel points \cite{paconv,kpconv,kpconvx,feng2020point}, leveraging key points as measurement references \cite{pointnat,pointdistribution}, or employing attention mechanisms for weight computation \cite{pct,pointtransformerv1,pointtransformerv2,spikingtransformer}.

However, these methods operate at point-level weight measurement, overlooking channel-wise sub-structure analysis and the influence of overall neighborhood distribution on individual point weight assignment.

To address these limitations, we propose a channel-wise weight calibration method that dynamically adjusts the contribution of neighboring points at channel granularity while fully considering the homogeneity of the overall neighborhood distribution, enabling adaptive fine-grained feature allocation based on the  neighborhood distribution.
\subsection{Channel Attentive}
To capture fine-grained features, channel attention mechanisms have gained extensive attention in computer vision. \cite{senet} identifies critical channels by computing average response weights; \cite{fcanet} enriches channel descriptors with higher-order statistics or frequency-domain features; \cite{dgcw} refines weight granularity by comparing channel differences per pixel pair; \cite{canet,overlapping,wang2020eca} leverage multi-head channel attention to refine neighborhood aggregation weights from spatial and semantic dimensions.

However, these methods consider channel responses from a global distribution perspective, overlooking local structural differences among points. Although \cite{dgcw,canet} attempt to address this, their computation still relies on global response differences along the channel dimension, measuring channel-wise weights for point-pair features rather than local similarity-based substructure correlations. They remain point-level weight computations relying on implicit modeling (attention/MLP), incurring high parameter overhead and low interpretability.

To address these limitations, we move beyond global channel distribution and instead introduce substructure correlation modeling based on local neighborhood distribution, refining channel-wise relevance weights for each point during neighborhood aggregation. Compared to globally shared channel weights, this neighborhood-adaptive approach offers greater flexibility and interpretability with low computational overhead.

\section{Preliminaries}
\subsection{Point Cloud Analysis Pipeline}
\label{sec:Point Cloud Analysis Pipeline}
This chapter briefly introduces existing point cloud analysis networks. We first introduce the commonly adopted hierarchical architecture: extracting features from local to global scales through multiple rounds of cross-resolution neighborhood aggregation. This paradigm was pioneered by PointNet++ \cite{pointnet++} and extensively extended since.

The PointNet++-style framework builds feature pyramids primarily via Set Abstraction (SA) modules:
\begin{align}
\label{eq:sa}
{f}_{i}^{\prime }=NA\left \{ p_i,f_i,\mathcal{N}_{i}^{\prime}\right\}
\end{align}
\begin{align}
\label{eq:na}
NA\left \{ p_i,f_i,\mathcal{N}_{i}^{\prime}\right\}=\mathcal{R}\left \{ \mathcal{M}\left \{ {f}_{i,j},{p}_{i}-{p}_{i,j}\right \}\mid j\in {\mathcal{N}}_{i}^\prime\right \}
\end{align}
where $p_i$ and $f_i$ denote point coordinates and features, and $\mathcal{N}_i$ denotes the neighborhood. The neighborhood $\mathcal{N}_{i}^{\prime}$ for a downsampled point $p_i \in N^\prime$ is gathered from the original point cloud $N$:
\begin{align}
\label{eq:ds}
\mathcal{N}_{i}^{\prime}=\mathcal{G}\left \{ p_i,p_j\right\},p_i \in N^\prime, p_j \in N
\end{align}

The SA module serves as the main component of a network layer. To enhance representation capacity, multiple SA modules with downsampling operations removed are typically stacked after the first SA module within the same layer to improve performance on complex tasks. This stacking operation enables feature propagation within local regions, thereby expanding the receptive field and enhancing non-linear representation capability:
\begin{align}
\label{eq:la}
{f}_{i}^{l+1}=EMBED(NA\left \{ p_i^l,f_i^l,\mathcal{N}_{i}\right\})
\end{align}
where $l$ denotes the module index, $\mathcal{N}_i$ is the neighborhood set from the current layer, and $\text{EMBED}(\cdot)$ is an additional encoding layer (e.g., MLP).

\begin{figure}[htbp]
    \centering
    
    \subfloat[{\small Existing point-level optimization strategies.}]{
        \includegraphics[width=0.49\textwidth]{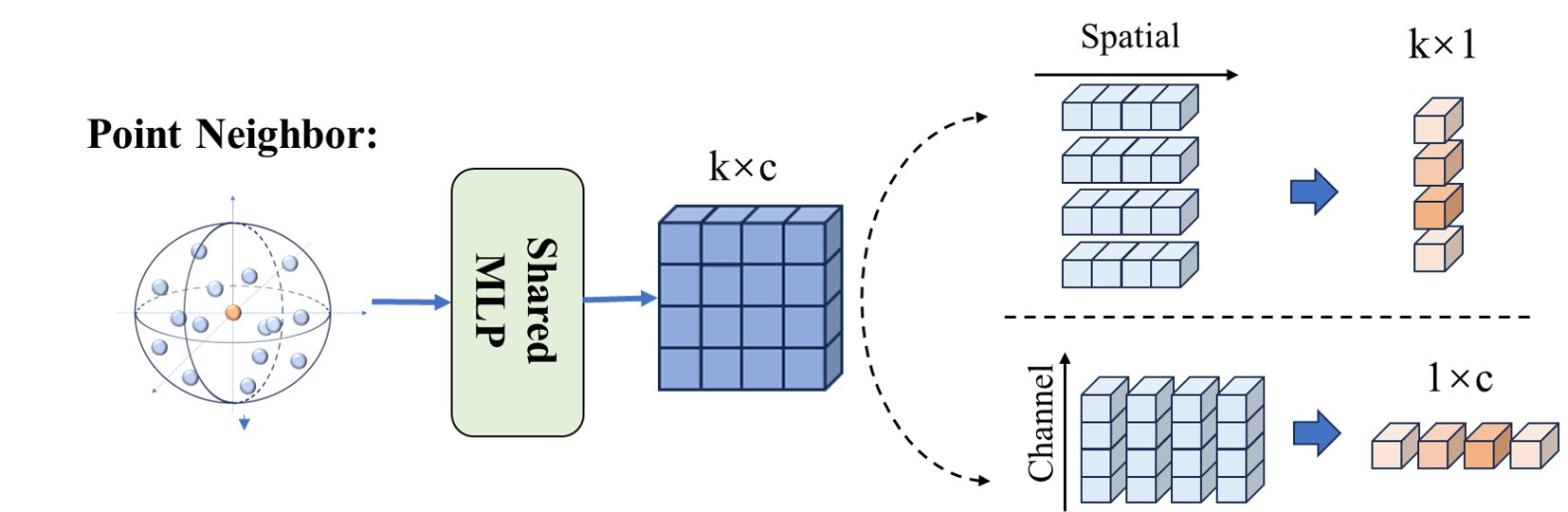}
        \label{fig:pointwise}
    }
    
    \vspace{0.3cm}
    
    \subfloat[{\small Proposed channel-level optimization strategies.}]{
        \includegraphics[width=0.49\textwidth]{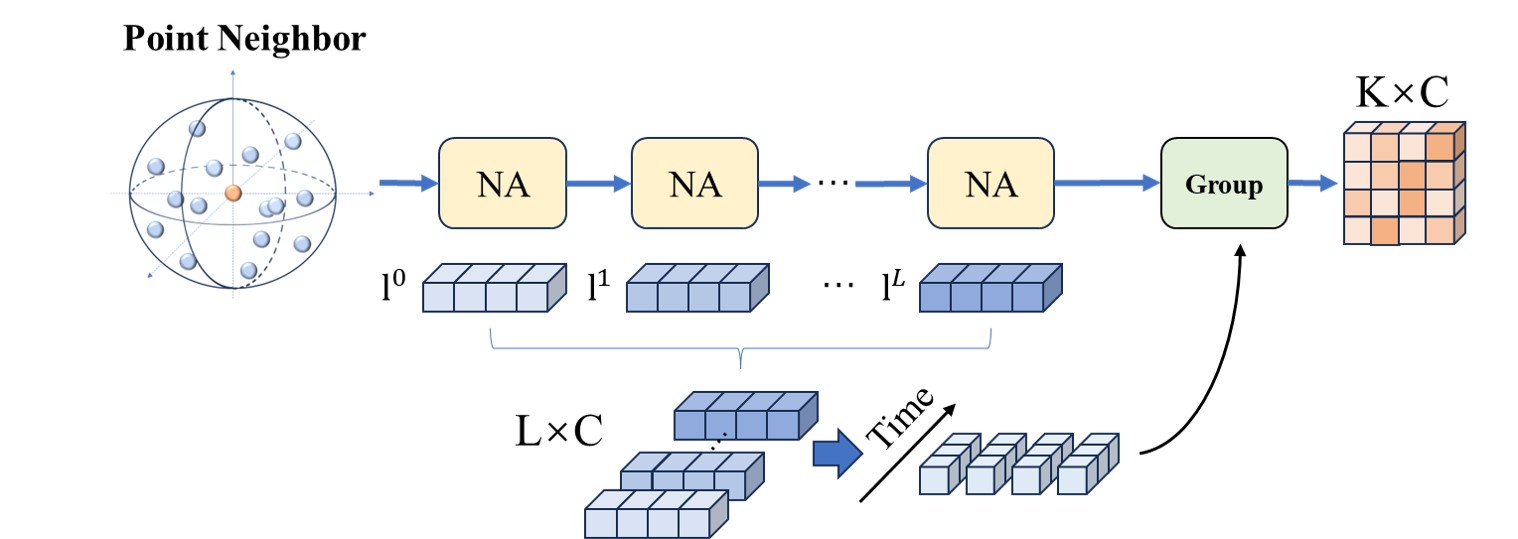}
        \label{fig:channelwise}
    }
    
    \vspace{0.2cm}

    \caption[Short caption for LoF]{{\small Illustration of optimization strategies in point cloud neighborhood aggregation ($K$: neighbor number, $C$: feature dimension, $L$: module number). (a) Existing methods: Upper (spatial attention) computes point-wise weights based on matrix rows; lower (channel attention) computes channel discriminability based on matrix columns. (b) Ours: computes element-wise correlations based on feature variation trends.}}
    \label{fig:method}
\end{figure}
\subsection{Limitations of Existing Methods}
\label{sec:Limitations of Existing Methods}
The architecture in Section \ref{sec:Point Cloud Analysis Pipeline} suffers from inherent deficiencies in its measurement mechanism. This section analyzes these problems and limitations of current approaches.

We analyze the SA module from coordinate and feature perspectives. In Eq. \ref{eq:sa}-\ref{eq:na}, the SA module takes a neighbor set collected by a Grouper, where coordinates describe geometry and features encode semantics. Due to shared MLP and batch normalization across all neighbors, each channel encodes identical semantic or geometric patterns across different points.

However, three problems exist: (i) Sampling (KNN/Ball Query) relies solely on geometric distance, inevitably introducing irrelevant points or noise. (ii) In high-level layers, each point represents a superpoint of stacked substructures; point-level similarity leads to substructure-level confusion. (iii) Shared MLP and coordinates' inability to measure semantic distance force max pooling to capture key features, but causing significant information loss. Thus, multi-layer networks require fine-grained substructure measurement and precise weight allocation for accurate aggregation.

Existing research follows two directions (Fig. \ref{fig:pointwise}): point cloud denoising networks \cite{x-3d,paconv,pointconvformer,pointdistribution} refine point-pair descriptions to suppress noise; channel attention mechanisms \cite{canet,overlapping,fcanet} identify sensitive channels via global statistics to enhance discriminability.

While these methods refine relation measurement, the former (optimizing rows) forces all channels to share a single point-level weight, causing structural confusion; the latter (optimizing columns) forces all neighbors to share the same channel weight, causing point-level confusion. Both remain confined to single-dimension optimization, failing to jointly model substructures and channel subspaces, thus struggling to capture fine-grained geometric-semantic correlations.

To address this, we propose a substructure-level joint optimization method for accurate feature propagation and noise suppression:
\begin{align}
\label{eq:con}
f_i^{\prime} = \big\Vert_{d=1}^{C} \left( \underset{j \in \mathcal{N}_i}{\mathrm{Mean}} \left( w_{ijd} \cdot c_{jd} \right) \right)
\end{align}
where $c_{jd}$ is the feature value of neighbor $j$ in channel $d$, $w_{ijd}$ is the calibration weight, and $\mathrm{Mean}$ denotes mean pooling. Each channel independently calibrates weights, enabling fine-grained channel-wise aggregation.

\section{Method}
\label{sec:method}
\subsection{Sequential Module Channel-wise Distance Computation}
\label{sec:sequential}
In Sec. \ref{sec:Limitations of Existing Methods}, we noted that existing methods address spatial relationships and channel discrimination separately but fail to achieve joint modeling. To address this, we introduce sequential module output trends as a new measurement basis and propose a channel-wise distance computation based on layer-wise feature transformation trends.

Building on the sequential module structure in existing networks, we introduce a temporal dimension and concatenate feature responses across channels by output sequence as the similarity measurement basis. As discussed in Sec. \ref{sec:Point Cloud Analysis Pipeline}, due to shared weights, each channel encodes consistent semantics, and module stacking promotes feature propagation and key homogeneous feature enhancement. Thus, similar structures exhibit similar response patterns. Comparing feature transformation trend consistency across channels between neighboring point pairs evaluates their structural similarity, avoiding extreme value effects from noise better than single-layer outputs. We achieve this by computing cosine angles of transformation vectors between sequential module outputs.

For center point $p_i$ and neighbor $p_j$, let $f_{i,d}^{(l)}$ be the feature of point $p_i$ in channel $d$ at the $l$-th module output ($l=1,...,L$). Define the transformation vector from layer $l$ to $l+1$ in channel $d$ as:
\begin{equation}
t_{i,d}^{(l)} = f_{i,d}^{(l+1)} - f_{i,d}^{(l)}, \quad l = 1,...,L-1
\end{equation}
$t_{i,d}^{(l)}$ reflects the direction and magnitude of feature changes between adjacent modules. The SA module promotes feature propagation within the neighborhood, leading to feature convergence and weight over-saturation. (Experimental details in Sec. \ref{sec: analysis study} and supplementary material (\textcolor{red}{Sec. S3})), we perform neighborhood normalization:
\begin{equation}
\label{eq:delta}
\Delta_{i,d}^{(l)} = t_{i,d}^{(l)} - \sum_{j=1}^{K} \frac{t_{j,d}^{(l)}}{K}, \quad l = 1,...,L-1
\end{equation}
where $t_{j,d}^{(l)}$ is the neighbor's trend value and $K$ is the number of neighbors. We use $\Delta_{i,d}^{(l)}$ for trend comparison, measuring directional consistency via cosine similarity:
\begin{equation}
\cos\theta_{ij,d}^{(l)} = \frac{\Delta_{i,d}^{(l)} \cdot \Delta_{j,d}^{(l)}}{|\Delta_{i,d}^{(l)}| |\Delta_{j,d}^{(l)}|}
\end{equation}
Accumulating over $L-1$ transformations yields the final trend similarity:
\begin{equation}
S_{ij,d} = \sum_{l=1}^{L-1} \frac{\cos\theta_{ij,d}^{(l)}}{(L-1)}
\end{equation}
Larger $S_{ij,d}$ indicates greater consistency in feature evolution trends and local geometric substructure similarity.

Note that we do not apply Softmax normalization, as it would weaken fine-grained discriminability.

To reduce computation, we introduce channel grouping ($G=4$, adjacent channels share similarity):
\begin{equation}
Pc_{ij,d} = S_{ij,g}, \quad g = \lceil d/G \rceil, g \in \{1,...,\lceil C/G \rceil\}
\end{equation}
This preserves statistical stability while reducing overhead (Experimental details in Sec. \ref{sec: analysis study}).

\subsection{Neighborhood Homogeneity-Guided Weight Calibration}
\label{sec:neighborhood homogeneity-guided weight calibration}
The initial channel-wise weights $Pc_{ij,d}$ obtained in Sec. \ref{sec:sequential} can preliminarily characterize feature similarity between neighboring points and the center point, but do not account for the influence of local neighborhood distribution on feature evolution.

The reference reliability of $Pc_{ij,d}$ is positively correlated with neighborhood homogeneity. In homogeneous regions (e.g., smooth surface interiors), neighboring points exhibit highly consistent local structures; as network depth increases, feature responses converge, enhancing correlations among similar points. In heterogeneous regions (e.g., edges, corners), neighboring points may belong to different geometric primitives, making feature response consistency difficult to maintain.

\begin{figure}[ht]
\centering
\includegraphics[width=0.48\textwidth]{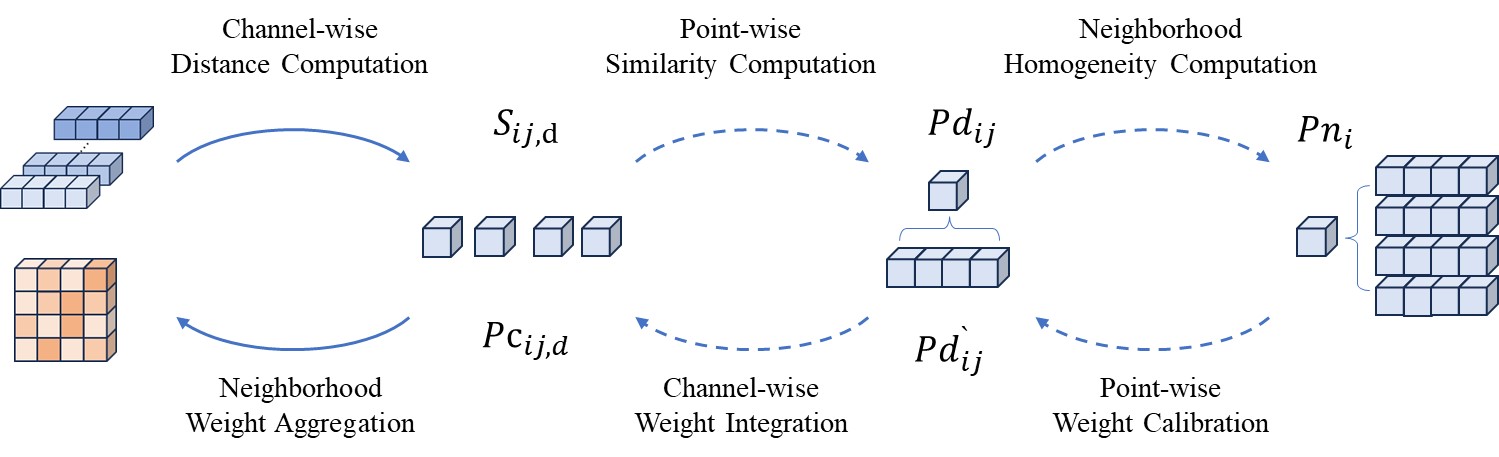}
\caption{Illustration of the three-level weight calibration framework guided by neighborhood distribution homogeneity. The framework takes initial channel-wise weights $Pc_{ij,d}$ as input and outputs calibrated weights $w_{ij,d}$.}
\label{fig:calibration_framework}
\end{figure}

This motivates weight calibration based on neighborhood homogeneity: amplify weight polarization in homogeneous regions to suppress noise; moderate weight differentiation in heterogeneous regions to preserve fine structures. We propose a three-level calibration framework that progressively quantifies local characteristics from channel-level $Pc$ to point-level $Pd$ to neighborhood-level $Pn$, and adaptively adjusts weight differentiation in a feedback manner. The overall pipeline is illustrated in Fig. \ref{fig:calibration_framework}.

First, aggregate channel-wise weights along the channel dimension to obtain point-wise similarity $Pd_{ij}$:
\begin{equation}
Pd_{ij} = \frac{1}{C}\sum_{d=1}^{C} Pc_{ij,d}
\label{eq:pd_aggregation}
\end{equation}

To quantify neighborhood homogeneity, analyze the distribution of point-wise similarities $Pd$ within the neighborhood. Let $v_i$ be the actual variance and $v_i^{\max}$ the theoretical maximum variance. Then $Pn_i$ is defined as:
\begin{equation}
Pn_i = 1.0- \exp\left(-\frac{v_i}{v_i^{\max}}\right)
\label{eq:pn_definition}
\end{equation}
Smaller $Pn_i$ values indicate more uniform distribution of $Pd$ within the neighborhood (homogeneous regions), while larger values indicate more dispersed distribution (edge/corner regions). The visualization results are provided in the \textbf{supplementary material} (\textcolor{red}{Sec. S3}).

Based on $Pn_i$, we adopt power-function scaling to calibrate point-wise similarity $Pd_{ij}\in[0,1]$:
\begin{equation}
Pd'_{ij} = (Pd_{ij} + \epsilon)^{\gamma(Pn_i)}
\label{eq:pd_calibration}
\end{equation}
\begin{equation}
\gamma(Pn_i) =  \exp\big(\alpha_n \cdot (\zeta - Pn_i )\big)
\label{eq:gamma}
\end{equation}
where $\zeta$ is the threshold for identifying homogeneous regions, $\alpha_n$ controls scaling strength, and $\epsilon$ is a small constant for numerical stability.

To enhance feature differentiation across channels, we introduce a learnable linear transformation on the initial channel weights:
\begin{equation}
Pc_{ij,g}^` = c \cdot \big( \sigma(a \cdot (Pc_{ij,g} - b)) - \sigma(-a \cdot b) \big)
\label{eq:pc_computation}
\end{equation}
where $a$, $b$, and $c$ are learnable parameters.

Finally, we combine the calibrated point-wise weight $Pd'_{ij}$, neighborhood-level indicator $Pn_i$, and transformed channel-wise weight $Pc'_{ij,d}$ to obtain the final weight:
\begin{equation}
w_{ij,d} = pd'_{ij} \cdot  Pc'_{ij,d} = (Pd_{ij} + \epsilon)^{\gamma(Pn_i)} \cdot Pc'_{ij,d}
\label{eq:final_weight}
\end{equation}

The overall procedure is summarized in \textbf{supplementary material} (\textcolor{red}{Sec. S1}). 
\subsection{Overall Architecture and Loss Function}
\begin{figure*}[ht]
\centering
\includegraphics[width=0.9\textwidth]{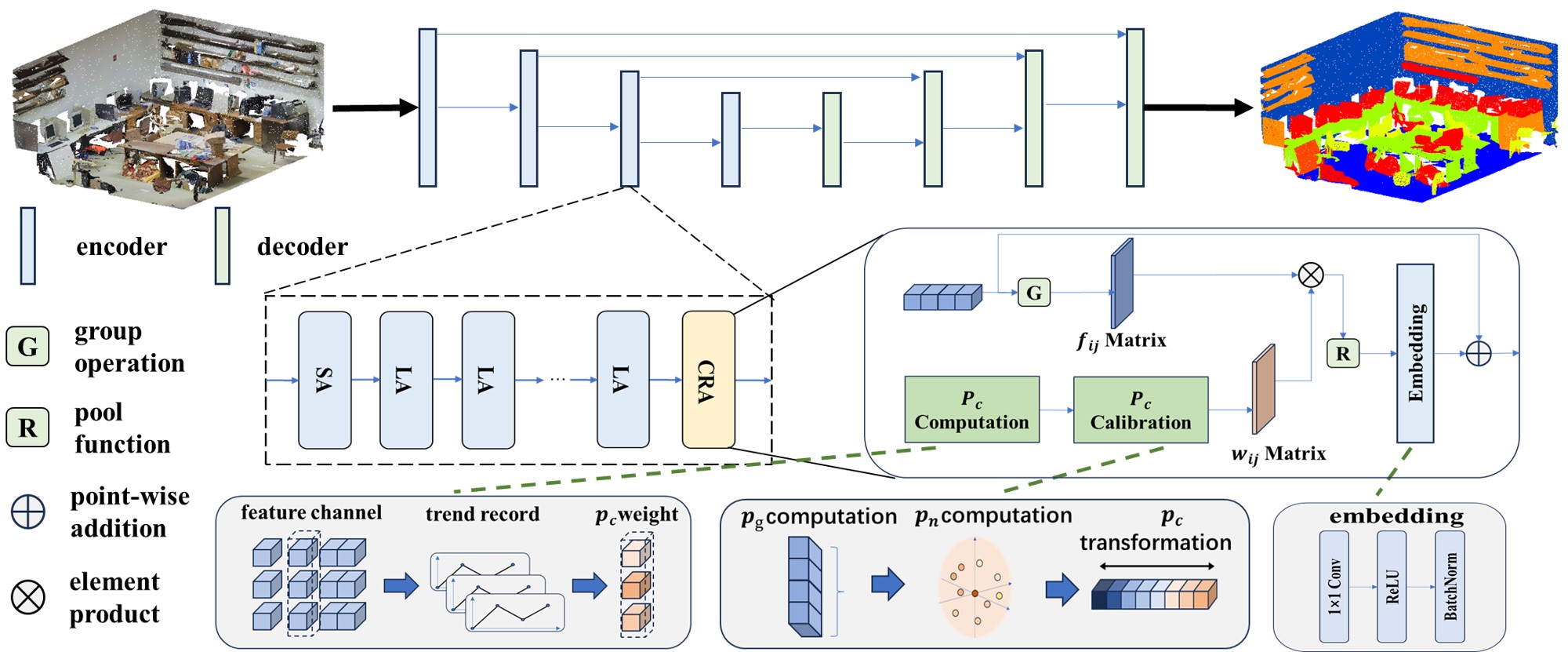}
\caption{Illustration of the mainstream point cloud analysis backbone architecture (top) and the deployment and internal structure of the proposed PointCRA (bottom). The mainstream architecture adopts an encoder-decoder framework, where only the encoder is used for classification tasks. PointCRA performs aggregation adjustment at the last layer of each encoding stage to compute channel-wise correlation weight based on the outputs of preceding serial modules. These weights are then applied to calibrate the features, followed by an embedding layer to align them with the original feature space, ensuring stable information flow to subsequent layers.}
\label{fig:cra_architecture}
\end{figure*}

Building upon the aforementioned analysis, this paper proposes a lightweight adaptive neighborhood aggregation network for point cloud analysis: PointCRA. The overall network architecture and the internal structure of the PointCRA are illustrated in Fig. \ref{fig:cra_architecture}.

The key of PointCRA lies in performing fine-grained aggregation correction at the end of each encoding stage. PointCRA retrieves the feature evolution sequence from preceding continuous neighborhood aggregation modules, capturing the response trajectory of each point across multiple aggregation steps. Based on this statistical information, it computes adaptive weights via the three-level calibration framework and calibrates the current layer's features. The calibrated features are then fed into an embedding layer with the same architecture as the backbone, where they are remapped to the existing feature space via standard neighborhood aggregation, completing the feature enhancement for that layer.

For the adaptive scaling of $Pc$ in Eq. \ref{eq:pc_computation}, we aim to enhance the discriminability across channels while avoiding extreme differentiation:
\begin{equation}
\mathcal{L}_{\text{reg}} = \mathbb{E}[\text{softplus}(b)] + \text{softplus}(1 - a) + \mathcal{P}(c; \phi_l, \phi_h)
\end{equation}

where $\phi_l$ and $\phi_h$ are the predefined parameter bounds. This loss function guides the formulation in Eq. \ref{eq:pc_computation} to smooth the initial $Pc$, preventing extreme binary differentiation and thereby enhancing the discriminative capability of the network.

To enhance the discriminability across channels, we introduce a loss function based on the Pearson correlation coefficient for the final weights $w_{ij,d}$. By minimizing the projection between different channel pairs, this loss encourages different channels to capture distinct and complementary information, thereby improving the representation capacity of the network. The formula is as follows:
\begin{equation}
\mathcal{L}_{\text{orth}} = \frac{1}{C(C-1)} \sum_{d_1 \neq d_2} \left| \frac{ \mathbf{w}_{d_1} \cdot \mathbf{w}_{d_2} }{ \| \mathbf{w}_{d_1} \|_2 \| \mathbf{w}_{d_2} \|_2 } \right|
\end{equation}
where $C$ is the total number of channels, $\mathbf{w}_{d} \in \mathbb{R}^{N \cdot K}$ denotes the weight vector of channel $d$ across all point pairs, and $\|\cdot\|_2$ represents the $\ell_2$ norm. The absolute value of the cosine similarity is used to penalize both positive and negative correlations.

The overall training objective of the network is defined as:

\begin{equation}
\mathcal{L} = \mathcal{L}_{task} + \lambda_1 \cdot \mathcal{L}_{reg} + \lambda_2 \cdot \mathcal{L}_{\text{orth}} 
\label{eq:total_loss}
\end{equation}

where $\mathcal{L}_{task}$ denotes the main task loss (e.g., cross-entropy loss for classification or Dice loss for segmentation), and $\lambda_1$ and $\lambda_2$ are weighting coefficient balancing the auxiliary loss.
\section{Experiments}
\label{sec:Experiment}
To demonstrate the effectiveness of our approach, we conduct experiments on three standard benchmarks for semantic segmentation and classification tasks, i.e., S3DIS \cite{s3dis} for segmentation, ShapeNetPart\cite{shapenet} for object part segmentation and ScanObjectNN \cite{scanobjectnn} for classification. We evaluate our method on different point cloud analysis backbones with stage-wise cascaded LA modules, aiming to comprehensively assess its generalization capability. All models are trained on a NVIDIA Geforce RTX 5090 32-GB GPU with a 20-core Intel Core i7-14700K CPU @3.40Ghz. We employ the CrossEntropy loss and optimize all models using the AdamW optimizer. 
\begin{table*}[ht]
\setlength{\tabcolsep}{2.8pt}
    \centering
    \caption{Semantic Segmentation Results on The S3DIS Benchmark.}
    \begin{tabular}{lccccccccccccccccc}
        \toprule
         Method & \rotatebox{90}{Params(M)} & \rotatebox{90}{OA(\%)} & \rotatebox{90}{mAcc(\%)} & \rotatebox{90}{mIoU(\%)} & \rotatebox{90}{ceiling} & \rotatebox{90}{floor} & \rotatebox{90}{wall} & \rotatebox{90}{beam} & \rotatebox{90}{column} & \rotatebox{90}{window} & \rotatebox{90}{door} & \rotatebox{90}{table} & \rotatebox{90}{chair} & \rotatebox{90}{sofa} & \rotatebox{90}{bookcase} & \rotatebox{90}{board} & \rotatebox{90}{clutter} \\
        \midrule
        PointNet\cite{pointnet} (CVPR 2017)& - & - & - & 41.1 & 88.8 & 97.3 & 69.8 & 0.1 & 3.9 & 46.3 & 10.8 & 59.0 & 52.6 & 5.9 & 40.3 & 26.4 & 33.2 \\
        
        PointNet++\cite{pointnet++} (NIPS 2017)& 1.0 & 83.0 & - & 53.5 & - & - & - & - & - & - & - & - & - & - & - & - & - \\

        PTV1\cite{pointtransformerv1} (CVPR 2021)& - & 90.8 & - & 70.4 & 94.0 & 98.5 & 86.3 & 0.0 & 38.0 & 63.4 & 74.3 & 89.1 & 82.4 & 74.3 & 80.2 & 76.0 & 59.3 \\
        
        PAGWN\cite{PAGWN} (IVC 2025)& - & 91.4 & 78.2 & 72.2 & 95.0 & 98.1 & 85.9 & 0.0 & 48.7 & 62.0 & 70.0 & 82.7 & 92.1 & 84.4 & 78.3 & 76.8 & 64.0 \\ 
        
        SAT\cite{zhou2023sat} (Arxiv 2023)& - & - & - & 72.6 & 93.6 & 98.5 & 87.2 & 0.0 & 49.3 & 61.1 & 73.6 & 83.7 & 91.8 & 81.7 & 77.9 & 82.3 & 63.4 \\ 
        
        HgCA\cite{hgca} (NC 2026)& - & 91.3 & 78.8 & 72.8 & 95.8 & 98.9 & 86.5 & 0.0 & 47.9 & 66.2 & 75.8 & 89.8 & 91.5 & 81.6 & 74.4 & 78.7 & 60.5 \\ 
        
        PointHR\cite{qiu2023pointhr} (Arxiv 2023)& - & 91.8 & - & 73.2 & 94.0 & 98.5 & 87.5 & 0.0 & 53.7 & 62.9 & 80.2 & 84.2 & 92.5 & 75.4 & 76.5 & 84.8 & 61.8 \\

        PointDistribution\cite{pointdistribution} (TVC 2025)& 28.8  & 91.8 & 79.2 & 73.4 & 95.7 & 98.6 & 85.7 &  0.0 &  43.7 & 57.9 & 78.4 & 85.3 & 92.6 & 88.0 &  78.9 & 83.5 & 65.9 \\
        
        PTV3\cite{pointtransformerv3} (CVPR 2024) & 46.2 & 91.7 & 78.9 & 73.4 & - & - & - & - & - & - & - & - & - & - & - & - & - \\

        CloudMamba\cite{cloudmamba} (AAAI 2026)& 16.6 & - & - & 73.6 & - & - & - & - & - & - & - & - & - & - & - & - & - \\
        
        DITR\cite{ditr} (3DV 2026) & - & - & - & 74.1 & - & - & - & - & - & - & - & - & - & - & - & - & - \\
        
        PointLearner\cite{pointlearner}(ICLR 2026) & - & - & - & 74.3 & - & - & - & - & - & - & - & - & - & - & - & - & - \\
        
        HSBNet\cite{hsbnet} (IF 2026) & - & - & - & 75.2 & - & - & - & - & - & - & - & - & - & - & - & - & - \\

        PTV3 + sonata\cite{sonata} (CVPR 2025) & 124.8 & 93.0 & 81.6 & 76.0 & - & - & - & - & - & - & - & - & - & - & - & - & - \\
        
        PointNext-L\cite{pointnext} (NIPS 2022)& 7.1 & 90.0 & 76.0 & 69.0 & 94.0 & 98.5 & 83.5 & 0.0 & 30.3 & 57.3 & 74.2 & 82.1 & 91.2 & 74.5 & 75.5 & 76.7 & 58.9 \\
        
        PointMetabase-L\cite{pointmeta} (CVPR 2023)& 2.7 & 90.5 & 76.0 & 69.5 & 94.6 & 98.3 & 84.8 & 0.0 & 31.9 & 59.1 & 75.8 & 80.7 & 90.8 & 76.6 & 77.3 & 74.0 & 62.0 \\

        DeLA-V1\cite{dela} (Arixv 2024)& 26.9 & 91.9 & 79.1 & 73.5 & 95.0 & 98.8 & 87.7 & 0.0 & 51.0 & 63.0 & 72.3 & 84.6 & 93.6 & 79.5 & 79.4 & 80.6 & 63.4 \\
        DeLA-V2\cite{dela}  (Arixv 2025)& 39.4 & 92.5 & 80.0 & 74.5 & 95.2 & 98.7 & 88.6 & 0.0 & 51.8 & 63.9 & 72.1 & 84.8 & 93.1 & 83.1 & 81.7 & 86.0 & 66.4 \\
        
        \midrule

        PointNext-L + CRA  & 7.5 & 90.3 (+0.3) & 76.2 (+0.2) & 70.0 (+1.0) & 94.8 & 98.5 & 83.4 & 0.0 & 36.6 & 60.3 & 73.6 & 81.9 & 90.2 & 76.4 & 76.4 & 78.3 & 59.2 \\
        PointMetaBase-L + CRA  & 3.1 & 90.6 (+0.1) & 76.1(+0.1) & 70.1 (+0.6) & 94.7 & 98.5 & 83.7 & 0.0 & 35.4 & 59.8 & 76.4 & 81.8 & 90.7 & 74.9 & 77.0 & 77.3 & 60.8 \\
        PointCRA (DeLA-V1+CRA) & 32.2 & 93.9 (+2.0) & 82.1 (+3.0) & 77.5 (+4.0) & 95.8 & 98.8 & 90.6 & 0.0 & 60.5 & 62.6 & 85.6 & 88.9 & 94.8 & 84.7 & 83.7 & 88.6 & 72.6 \\
        DeLA-V2 + CRA  & 44.8 & 93.6 (+1.1) & 82.2 (+2.2) & 77.3 (+2.8) & 95.8 & 98.8 & 89.8 & 0.0 & 63.3 & 65.7 & 83.3 & 87.2 & 92.5 & 85.5 & 82.6 & 89.8 & 70.6 \\
        \bottomrule   
    \end{tabular}
    \label{tab:s3dis}
\end{table*}

We design PointCRA based on DeLAv1 as the baseline, and select other baseline models (PointNext, PointMetaBase, DeLAv2) for extended experiments to validate the effectiveness and transferability of the proposed method. For the sake of brevity, we denote the proposed method as CRA. PointNext is a modernized variant of PointNet++ that achieves significant performance gains through advanced data augmentation strategies and improved network architectures. PointMetaBase builds upon the PointNeXt framework and introduces an explicit spatial encoding scheme to enhance the representation capability of the network. DeLA‑V1 realizes efficient and accurate spatial structure encoding by decoupling neighborhood aggregation from feature encoding and incorporating KNN-based edge pooling. DeLA‑V2 further improves upon DeLA‑V1 by refining the network architecture and introducing novel nonlinear transformations in the decoupled neighborhood aggregation modules to enhance representational capacity. To ensure a fair comparison, we select backbone versions with multi-stage cascaded architectures for which public code is available: all four backbones on S3DIS, the DeLA series (V1 and V2) on ScanNetV2 and DeLAv1 on ShapeNetPart.

To ensure a fair comparison, we keep the default settings and data processing methods of the original backbone network, and also maintain consistent loss function parameter configurations across all experiments: the homogeneous threshold is set to 0.7; the learnable weights $\alpha_d$ and $\beta_d$ for $Pc_{ij}$ and $d$ are initialized to $1$ and $0$, respectively; For the auxiliary regularization loss, the boundary thresholds are set to $\phi_l = 0.2$, $\phi_h = 0.8$ for parameter $c$, ensuring they remain within reasonable ranges during training.

For evaluation methods, we refer to previous work, using mean intersection over union (mIoU) and overall accuracy (OA) for semantic segmentation task and OA and mean accuracy (mAcc) for classification task.
\subsection{Semantic Segmentation on S3DIS}
\textbf{Dataset:} S3DIS\cite{s3dis} (Stanford Large-Scale 3D Indoor Spaces) is a large-scale indoor benchmark for point cloud semantic segmentation, reconstructed from RGB-D images captured by cameras equipped with structured light sensors. It comprises 3D point clouds captured from six large-scale indoor areas across three different buildings, covering over 6,000 square meters with 271 rooms . The dataset includes diverse architectural styles and functional spaces such as offices, conference rooms, hallways, restrooms, and lobbies . Each point is annotated with one of 13 semantic categories, including structural elements (ceiling, floor, wall, beam, column, window, door) and common furniture items (table, chair, sofa, bookcase, board). In our experiments, we evaluate on Area 5, which is widely adopted as a challenging test split due to its distinct scene distribution.

\textbf{Setup:}For experiments on all four backbone versions, the unified settings as follows: the input point cloud is downsampled with a voxel size of 0.4 m; the initial learning rate is set to 0.01, and the weight decay is set to $10^{-4}$; the batch size is set to 8. For PointNext and PointMetaBase, we fix the input to 24,000 points per sample, use an initial learning rate of $1\times10^{-2}$, and train for 100 epochs with cosine decay to $1\times10^{-4}$. For DeLA‑V1 and DeLA‑V2, we use a maximum of 30,000 input points, adopt an initial learning rate of $6\times10^{-3}$, and train for 110 epochs with decay to $6\times10^{-7}$.

\textbf{Result:}We evaluate our model using the best model of validation set to test the entire scene of S3DIS Area5, and the results are shown in Tab.\ref{tab:s3dis}. As shown in the bottom section of Table~\ref{tab:s3dis}, incorporating the CRA consistently improves the performance across all baseline models. Specifically, $PointNeXt-L + CRA$ achieves 70.0\% mIoU (↑1.0), $PointMetaBase-L + CRA$ attains 70.1\% mIoU (↑0.6), $DeLA-V2 + CRA$ reaches 77.3\% mIoU (↑2.8), and $PointCRA$ ($DeLA-V1 + CRA$) achieves state-of-the-art level performance:  93.9\% OA (↑2.0) 82.1\% mAcc (↑3.0), 77.5\% mIoU (↑4.0).

The specific visualization results of all four backbones are shown in \textbf{supplementary material} (\textcolor{red}{Sec. S2}).

Experimental results demonstrate that the four backbone networks, despite their diverse training frameworks and feature encoding paradigms, achieve consistent performance gains after incorporating the CRA method. The improvements across subcategories exhibit a similar consistency, with particularly notable gains on challenging classes that are prone to confusion with adjacent objects due to analogous spatial distributions, such as ceiling, window, door, and clutter.

\subsection{Classification on ScanObjectNN}
\textbf{Dataset:} ScanObjectNN\cite{scanobjectnn} is a real-world 3D point cloud dataset for object classification, introduced to address the limitations of synthetic datasets like ModelNet40. The dataset is derived from the SceneNN dataset, which collects real indoor scenes using an RGB-D sensor (Kinect v2) . Objects are automatically segmented and extracted from these reconstructed indoor scenes, resulting in approximately 15,000 objects across 15 categories, sourced from 2,902 unique instances . The dataset provides multiple variants with increasing difficulty to simulate real-world challenges such as background clutter, occlusion, and object partiality. In our experiments, we adopt the PB\_T50\_RS variant, which is the most challenging and widely used benchmark setting. It incorporates random translation, random rotation, and uniform scaling to jointly simulate perturbation, background, and scaling effects.

\begin{table}[ht]
\setlength{\tabcolsep}{4pt}
    \centering
    \caption{Classification Results On The ScanObjectNN Benchmark.}
    \begin{tabular}{lccc}
        \toprule
        Method & OA(\%) & mAcc(\%) & Param (M) \\
        \midrule
        PointNet\cite{pointnet} (CVPR 2017)& 68.2 & 63.4 & 3.5 \\
        DGCNN\cite{chen2023ddgcn} (TVC 2023) &  78.1  & 73.6 & 1.8 \\
        PointCNN\cite{pointcnn} (NIPS 2018)& 78.5 & 75.1 & 0.6 \\
        PointStack\cite{pointstack} (RS 2024) & 86.9 & 85.8 & - \\
        DGCNN(ACN)\cite{putra2025adacrossnet} (IJIES 2025) &  82.1 & - & - \\
        PCM\cite{pcm} (AAAI 2025)& 88.1 & 86.6 & 34.2 \\
        PGCF\cite{xu2026pgcf} (PR 2026)& 89.3 & 88.2 & 1.4 \\
        MVFormer\cite{wang2026mvformer} (IJCV 2026)&  90.7 & - & - \\
        CPG\cite{zhou2026cpg} (PR 2026)& 91.3 & - & - \\
        DeLAv1 \cite{dela} (Arxiv 2024)& 90.1 & 88.6 & 5.3 \\
        DeLAv2 \cite{dela} (Arxiv 2025)& 91.3 & 90.3 & 3.5 \\
        \midrule
        PointCRA (DeLAv1+CRA) & 90.4(+0.3) & 89.3(+0.7) & 6.1 \\
        DeLAv2+CRA & 91.6(+0.3) & 90.6(+1.3) & 4.3 \\
        \bottomrule
    \end{tabular}
    \label{tab:classification}
\end{table}

\textbf{Setup:} The unified settings as follows: we apply random rotation around the Y-axis, random scaling in [0.9, 1.1], point shuffling, and the fix the input to 2048 points per sample. For DeLA series experiments, we follow the original settings with a batch size of 32, label smoothing of 0.2, initial learning rate of $3\times10^{-3}$, and decay rate of $5\times10^{-2}$. DeLA-V1 is trained for 250 epochs, while DeLA-V2 is trained for 400 epochs.

\textbf{Result:} We evaluate our model on the most challenging PB\_T50\_RS variant of ScanObjectNN, and the results are shown in Tab.\ref{tab:classification}. As shown in the bottom section of Table~\ref{tab:classification}, incorporating the CRA consistently improves performance across both DeLA backbones. Specifically, PointCRA($DeLAv1 + CRA$) achieves 90.4\% OA (↑0.3) and 89.3\% mAcc (↑0.7), while $DeLAv2 + CRA$ achieves 91.6\% OA (↑0.3) and 90.6\% mAcc (↑1.3). Both experimental versions achieved similar improvements after deploying CRA.

\subsection{Object Part Segmentation on ShapeNetPart}
\textbf{Dataset:} ShapeNetPart\cite{shapenet} is a widely used 3D point cloud dataset for fine-grained part segmentation, derived from ShapeNetCore, a large-scale repository of 3D CAD models. The dataset comprises 16,881 models across 16 object categories, with each category containing 2 to 6 parts, resulting in a total of 50 annotated part segments. Each point in the point cloud is annotated with a part category label (e.g., airplane wings, table legs, chair arms). Following the standard evaluation protocol, we train on the official training split (14,006 models) and report instance-average mIoU and class-average mIoU on the test split (2,874 models).

\textbf{Setup:} For PointCRA experiment on ShapeNetPart, we adopt the following configuration: The input points are normalized to a range of 40 and each sample contains 2,048 points with normals; During training,the initial learning rate is set as $2\times10^{-3}$ with the weight decay of $5\times10^{-2}$. Label smoothing of 0.2 is applied during training. We train the model for 250 epochs with a batch size of 32.
\begin{table}[htbp]
\setlength{\tabcolsep}{2.5pt}
    \centering
    \caption{Classification Results On The ShapeNetPart Benchmark.}
    \begin{tabular}{lccc}
        \toprule
        Method & Ins mIoU(\%) & Cat mIoU(\%) & Param (M) \\
        \midrule
        PointNet++\cite{pointnet++} (NIPS 2017)& 82.1 & - & - \\
        S3DNet\cite{s3dnet} (NN 2025)& 85.0 & - & - \\
        AdaCrossNet\cite{putra2025adacrossnet} (IJIES 2025)&- &85.1& -\\
        DGCNN\cite{chen2023ddgcn} (TVC 2023) &  85.2  & - & - \\
        PointGL\cite{pointgl} (TMM 2024) &  85.6  & 83.8 & - \\
        PointCNN\cite{pointcnn} (NIPS 2018)& 86.1 & 84.6 & - \\
        PGCF\cite{xu2026pgcf} (PR 2026)& 87.2 & 85.5 & 22.7 \\
        CloudMamba\cite{cloudmamba} (AAAI 2026)& 86.6 & - & 16.6 \\
        MVFormer-l\cite{wang2026mvformer} (IJCV 2026)& 87.1 & - & - \\
        DeLAv1 \cite{dela} (Arixiv 2024)& 86.9 & 85.4 & 7.6 \\
        \midrule
        PointCRA & 87.4(+0.5) & 85.5(+0.1) & 9.2 \\
        \bottomrule
    \end{tabular}
    \label{tab:shapenetpart}
\end{table}

\textbf{Result:} The results are shown in Tab.\ref{tab:shapenetpart}. PointCRA achieves a 0.5\% improvement in instance mIoU over the baseline DeLAv1. Compared to other methods, PointCRA delivers competitive performance with low parameter overhead.


\subsection{Ablation Study}
\label{sec:ablation}
\textbf{Ablation of Main Improvements:} In order to further verify the effectiveness of PointCRA,  we choose the data set S3DIS Area5 for ablation study. For fair comparison, we did not change the training parameters.

We decompose the proposed PointCRA into three components: channel-wise weight $Pc$ calibration, a multi-level calibration framework based on neighborhood homogeneity distribution, and a learnable weight mapping. Accordingly, we design three ablation experiments: Experiment A introduces only the channel-wise $Pc$ calibration aggregation based on DeLA-V1; Experiment B incorporates three-level calibration upon experiment A; Experiment C further introduces a learnable scaling mapping upon experiment B. Experiment D further introduces the loss constraint upon experiment C. The experimental results are presented in Table~\ref{tab:ablation of CRA module}. The results from the four experimental groups validate the effectiveness of the proposed improvements.
\begin{table}[ht]
\setlength{\tabcolsep}{10pt}
  \centering
    \caption{Ablation Results of PointCRA.}
  \begin{tabular}{lcc}
    \toprule
Ablate & mIoU  & $\triangle$  \\
    \midrule
baseline(DeLAv1)& 73.5 &  - \\
\hline
$A:~\rightarrow Pc~~calibration$ & 76.2 &  +2.7 \\
$B:A~\rightarrow + three-level~calibration$ & 77.1 &  +0.9\\
$C:~B\rightarrow + learnable~scaling~ mapping$ & 77.3 & +0.2  \\
$D:~C\rightarrow + loss~constraint$ & 77.5 & +0.2  \\
    \bottomrule
  \end{tabular}
  \label{tab:ablation of CRA module}
\end{table}

\textbf{Ablation of Calibration:} To further validate the proposed three-level calibration framework, we conduct ablation experiments on S3DIS Area 5 using DeLAv1 as the baseline, keeping all training parameters consistent.

We decompose the weight calibration framework into four components, corresponding to four experiments: A: the basic weighted computation of channel-level weights \( P_c \); B: introducing the computation and weighting of point-wise similarities \( P_g \) based on A; C: introducing the computation of \( P_n \) and the calibration of \( P_g \) based on B; D: introducing a learnable scaling mapping for \( P_c \) based on C. The experimental results are presented in Table~\ref{tab:ablation of Calibration}.
The results from the four experimental groups validate the effectiveness of the calibration mechanism.
\begin{table}[ht]
\setlength{\tabcolsep}{10pt}
  \centering
    \caption{Ablation Results of Calibration.}
  \begin{tabular}{lcc}
    \toprule
Ablate & mIoU  & $\triangle$  \\
    \midrule
baseline(DeLAv1)& 73.5 &  - \\
\hline
$A:~\rightarrow Pc~~weighted$ & 76.2 &  +2.7 \\
$B:A~\rightarrow + Pg~weighted$ & 76.8 &  +0.6\\
$C:~B\rightarrow + Pn~computed~and~Pg~Calibrated$ & 77.1 & +0.3  \\
$D:~C\rightarrow + learnable~scaling~mapping$ & 77.3 & +0.2  \\
    \bottomrule
  \end{tabular}
  \label{tab:ablation of Calibration}
\end{table}

\subsection{Analysis Study}
\label{sec: analysis study}
\begin{figure}[ht]
\centering
\includegraphics[width=0.48\textwidth]{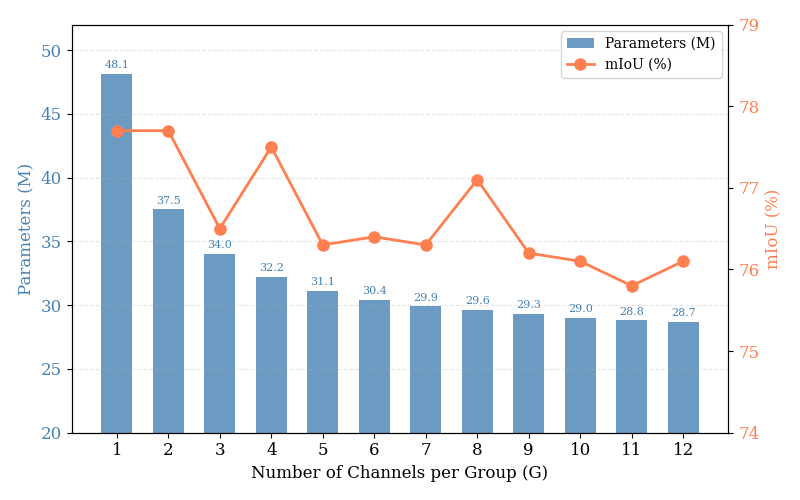}
\caption{Effect of Channel Group Size (G) on Performance (PointCRA).}
\label{fig:ablation of g}
\end{figure}

\textbf{Analysis of Group Size:} We also conduct comparative experiments on the number of feature channel groups $G$. The value of $G$ is varied from 1 to 12, and the model remains PointCRA on S3DIS Area5. For different settings of $G$, we adopt a zero-padding strategy to ensure that the number of feature channels can be completely grouped. The experimental results are shown in Figure~\ref{fig:ablation of g}. 

We can observe that as $G$ increases, the total number of parameters decreases correspondingly, following an exponential decay trend. Meanwhile, the overall performance remains at a comparable level in the early stage when $G = 1$ to $4$, and gradually declines thereafter. In terms of overall experimental results, if the value of $G$ is not divisible by the number of channels, it will have a significant negative impact. A smaller grouping granularity (i.e., a smaller $G$) leads to better performance but also incurs a larger parameter count. After comprehensively comparing the overall results, we select $G = 4$ as a balanced setting between performance and parameter overhead.

\begin{figure}[!htb]
    \centering
    
    \begin{subfigure}[b]{0.47\columnwidth}
        \centering
        \includegraphics[width=\textwidth]{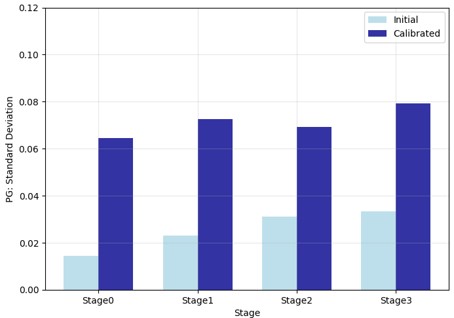}
        \caption{$Pg$ Standard Deviation}
        \label{fig:pgstd}
    \end{subfigure}\hfill
    \begin{subfigure}[b]{0.47\columnwidth}
        \centering
        \includegraphics[width=\textwidth]{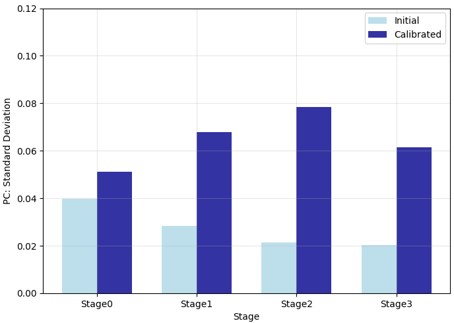}
        \caption{$Pc$ Standard Deviation}
        \label{fig:pcstd}
    \end{subfigure}
    
    \vspace{0.3cm}
    
    \begin{subfigure}[b]{0.47\columnwidth}
        \centering
        \includegraphics[width=\textwidth]{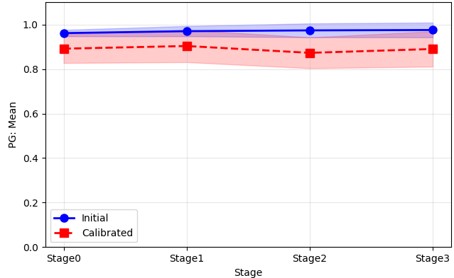}
        \caption{$Pg$ Mean}
        \label{fig:pgmean}
    \end{subfigure}\hfill
    \begin{subfigure}[b]{0.47\columnwidth}
        \centering
        \includegraphics[width=\textwidth]{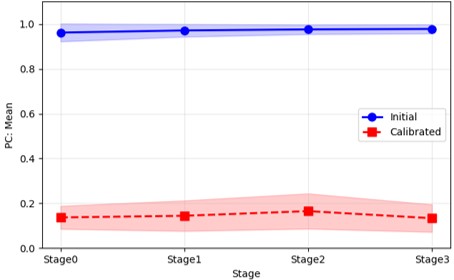}
        \caption{$Pc$ Mean}
        \label{fig:pcmean}
    \end{subfigure}
    
    \caption{Analysis of calibration}
    \label{fig:analysis of calibration}
\end{figure}

\textbf{Analysis of Weight Calibration:} To further analyze the effect of our proposed three-level weight calibration framework based on neighborhood distribution homogeneity, we conducted additional statistical experiments on the ShapeNetPart dataset. As illustrated in the visualization of $Pn$ distribution in \textbf{supplementary material} (\textcolor{red}{Sec. S3}), the homogeneity of point cloud features is progressively strengthened with increasing network depth, and point features tend to become increasingly similar. Accordingly, we collected the initial and calibrated weights of $Pg$ and $Pc$ across different stages. The statistical results are presented in Fig.~\ref{fig:analysis of calibration}.

As shown in Fig.~\ref{fig:pgstd} and \ref{fig:pcstd}, the standard deviation of both $Pg$ and $Pc$ weights increases notably after calibration, indicating enhanced feature discriminability. Furthermore, the mean distributions in Fig.~\ref{fig:pgmean} and \ref{fig:pcmean} demonstrate that the calibrated weights exhibit a broader distribution than the initially overly similar ones. This alleviates weight saturation and promotes better feature discrimination, leading to improved network performance.

\section{Conclusions}
This paper proposes PointCRA, a lightweight method for improving point cloud analysis networks via enhanced neighborhood aggregation. Unlike existing methods relying on coarse-grained point-level or channel-level metrics, PointCRA introduces a fine-grained channel-wise affinity metric based on feature transformation trends across sequential modules. Building on this, we construct a three-level calibration framework guided by neighborhood homogeneity for adaptive weight calibration, complemented by a dedicated loss function that enforces orthogonality constraints to enhance channel discriminability.

Extensive experiments on three public benchmarks demonstrate that integrating CRA consistently improves performance across various training frameworks and feature encoding paradigms. Ablation and visualization studies validate each component, while channel grouping analysis provides practical deployment guidance. Our method offers a lightweight, interpretable, and efficient solution for discriminative feature learning in point cloud analysis. Future work will focus on improving computational efficiency and evaluation methodology for fine-grained measurement in large-scale scenarios.

\bibliographystyle{IEEEtran}
\bibliography{main}

\section{Biography Section}
\vspace{-20pt}
\begin{IEEEbiographynophoto}{Jiaqi Shi}
He is currently pursuing the Ph.D. degree with the School of Automation Science and Electrical Engineering at Beihang University. His research interests include computer vision, especially point cloud analysis and three-dimensional scene perception.
\end{IEEEbiographynophoto}%
\vspace{-20pt}
\begin{IEEEbiographynophoto}{Jin Xiao}
She is a professor at the School of Automation Science and Electrical Engineering, Beihang University, Beijing, China. Her current research interests mainly include complex networks, pattern recognition, and embedded test systems.
\end{IEEEbiographynophoto}%
\vspace{-20pt}
\begin{IEEEbiographynophoto}{Xiaoguang Hu}
She is currently a Professor with the School of Automation Science and Electrical Engineering, Beihang University. Her research interests include swarm intelligence, embedded test systems, and smart grids.
\end{IEEEbiographynophoto}%
\vspace{-20pt}
\begin{IEEEbiographynophoto}{Wenxuan Ji}
He began pursuing the Ph.D. degree at Beihang University in 2023, where he is currently a third-year Ph.D. student. His research interests include 3D reconstruction and deep learning-based SLAM.
\end{IEEEbiographynophoto}%
\vspace{-20pt}
\begin{IEEEbiographynophoto}{Zichong Jia}
He is currently working toward the Ph.D. degree with the School of Automation Science and Electrical Engineering, Beihang University. His research interests include robot manipulation and computer vision.
\end{IEEEbiographynophoto}%
\vspace{-20pt}
\begin{IEEEbiographynophoto}{Zifan Long}
He is currently pursuing the master's degree with the School of Automation Science and Electrical Engineering at Beihang University. His research interests include computer vision, especially point cloud analysis.
\end{IEEEbiographynophoto}%
\vspace{-20pt}
\begin{IEEEbiographynophoto}{Tianyou Chen}
received the Ph.D. degree from Beihang University in 2024. He is now an engineer at Wuhan Leaddo Measuring \& Control Technology. His current research interests include computer vision and deep learning, especially instance segmentation and object detection.
\end{IEEEbiographynophoto}%
\vspace{-20pt}
\begin{IEEEbiographynophoto}{Baochang Zhang}
He is currently a Professor with Beihang University, Beijing, China. His current research interests include pattern recognition, machine learning, face recognition, and wavelets.
\end{IEEEbiographynophoto}%

\newpage

\vfill

\end{document}